\pdfoutput=1

\documentclass[11pt]{article}

\usepackage{acl}

\usepackage{times}
\usepackage{latexsym}
\usepackage{times}
\usepackage{latexsym}
\usepackage{booktabs}
 \usepackage{amsmath}
\usepackage{amssymb}
\usepackage{multirow}
\usepackage{subcaption}
\usepackage{makecell}
\usepackage{mathrsfs}
\usepackage{tabularx}
\usepackage{nicematrix}
\usepackage{threeparttable}
\usepackage{microtype}
\usepackage{bm}
\usepackage{amsfonts}
\usepackage{diagbox}
\usepackage{amsmath}
\usepackage{multirow}
\usepackage{graphicx}
\usepackage{mathtools}

\usepackage[T1]{fontenc}

\usepackage[utf8]{inputenc}

\usepackage{microtype}

\usepackage{inconsolata}

\usepackage{graphicx}

\title{RUPBench: Benchmarking Reasoning Under Perturbations \\ for Robustness Evaluation in Large Language Models}

\author{Yuqing Wang\thanks{Denotes equal contribution.} \\
  Stanford University \\
  \texttt{ywang216@stanford.edu} \\\And
  Yun Zhao\footnotemark[1] \\
  Meta Platforms, Inc. \\
  \texttt{yunzhao20@meta.com} \\}

\begin{document}
\maketitle
\begin{abstract}
With the increasing use of large language models (LLMs), ensuring reliable performance in diverse, real-world environments is essential. Despite their remarkable achievements, LLMs often struggle with adversarial inputs, significantly impacting their effectiveness in practical applications. To systematically understand the robustness of LLMs, we present RUPBench, a comprehensive benchmark designed to evaluate LLM robustness across diverse reasoning tasks. Our benchmark incorporates 15 reasoning datasets, categorized into commonsense, arithmetic, logical, and knowledge-intensive reasoning, and introduces nine types of textual perturbations at lexical, syntactic, and semantic levels. By examining the performance of state-of-the-art LLMs such as GPT-4o, Llama3, Phi-3, and Gemma on both original and perturbed datasets, we provide a detailed analysis of their robustness and error patterns. Our findings highlight that larger models tend to exhibit greater robustness to perturbations. Additionally, common error types are identified through manual inspection, revealing specific challenges faced by LLMs in different reasoning contexts. This work provides insights into areas where LLMs need further improvement to handle diverse and noisy inputs effectively. Our data and code are available at \url{https://github.com/EternityYW/RUPBench}.
\end{abstract}

\section{Introduction}
Large language models (LLMs) have gained increasing popularity due to their unprecedented performance in various tasks such as sentiment analysis~\citep{miah2024multimodal}, complex reasoning~\citep{wang2023can}, and time series analysis~\citep{zhao2021empirical, wang2022enhancing}. Models like GPT-3~\citep{brown2020language}, GPT-4o~\citep{gpt-4o}, and Llama3~\citep{llama3modelcard} have set new benchmarks in natural language processing, pushing the boundaries of what these systems can achieve. However, as the deployment of LLMs in real-world applications grows, particularly in high-risk domains, ensuring their robustness against diverse and potentially adversarial inputs becomes critical. Despite advancements, LLMs remain vulnerable to perturbations that can significantly degrade their performance. These perturbations can come in various forms, including lexical variations (e.g., typos), syntactic changes (e.g., cleft constructions), and semantic distractions (e.g., red herrings). Such weaknesses pose serious challenges, especially in applications requiring high reliability and accuracy, such as healthcare~\citep{wang2024fairehr}, legal document analysis~\citep{cheong2024not}, and automated customer service~\citep{kolasani2023optimizing}. 

Several studies have explored the robustness of LLMs from various angles. For instance, datasets like AdvGLUE~\citep{wang2021adversarial} and AdvGLUE++~\citep{wang2024decodingtrust} are specifically designed to test how language models respond to adversarial inputs, which are meticulously altered to elicit incorrect responses from the models. Wang et al.~\citep{wang2023robustness} assessed the robustness of ChatGPT and other LLMs against adversarial and out-of-distribution (OOD) samples, while Zhuo et al.~\citep{zhuo2023robustness} evaluated the robustness of semantic parsing. However, these studies focus on restricted tasks or types of perturbations, lacking a holistic evaluation framework that comprehensively assesses robustness across multiple categories and distinct perturbation types. Additionally, they do not delve deeply into the specific error patterns induced by different perturbations, leaving gaps in understanding how to enhance the models' resilience in practical applications.

To address this gap, we introduce the \textbf{R}easoning \textbf{U}nder \textbf{P}erturbations \textbf{Bench}mark (RUPBench), a comprehensive benchmark designed to evaluate the robustness of LLMs across different reasoning tasks. RUPBench includes 15 source datasets spanning four major reasoning categories: commonsense, arithmetic, logical, and knowledge-intensive. Each dataset is subjected to nine types of textual perturbations, covering lexical, syntactic, and semantic levels, to simulate real-world input variations. Then, we conduct extensive experiments with several leading LLMs using RUPBench, including GPT-4o~\citep{gpt-4o}, Llama3~\citep{llama3modelcard}, Phi-3~\citep{abdin2024phi}, and Gemma~\citep{team2024gemma} models, assessing their performance on both original and perturbed datasets. By analyzing the models' responses, we provide insights into their robustness and identify common error patterns. Our findings indicate that larger models generally exhibit greater robustness to perturbations. Manual inspection of incorrect predictions highlights specific error types prevalent across all LLMs, directing areas for improvement and emphasizing the need for targeted strategies to address these weaknesses by task.

In summary, our contributions are threefold:
\begin{enumerate}
\item[(1)] We introduce RUPBench, a comprehensive benchmark designed to systematically evaluate the robustness of LLMs across 15 reasoning tasks, incorporating nine types of textual perturbations, resulting in a total of 365,580 perturbed samples.
\item[(2)] We assess the performance of several state-of-the-art LLMs, including GPT-4o, Llama3, Phi-3, and Gemma, on both original and perturbed datasets. Our extensive analysis provides detailed insights into their robustness across different tasks and perturbations.
\item[(3)] We identify common error types from perturbations through manual inspection, highlighting challenges LLMs face, such as context misinterpretation and knowledge gaps, to guide future research towards more resilient and reliable LLMs.
\end{enumerate}

\begin{figure*}[htbp]
\centering
\includegraphics[width=\textwidth]{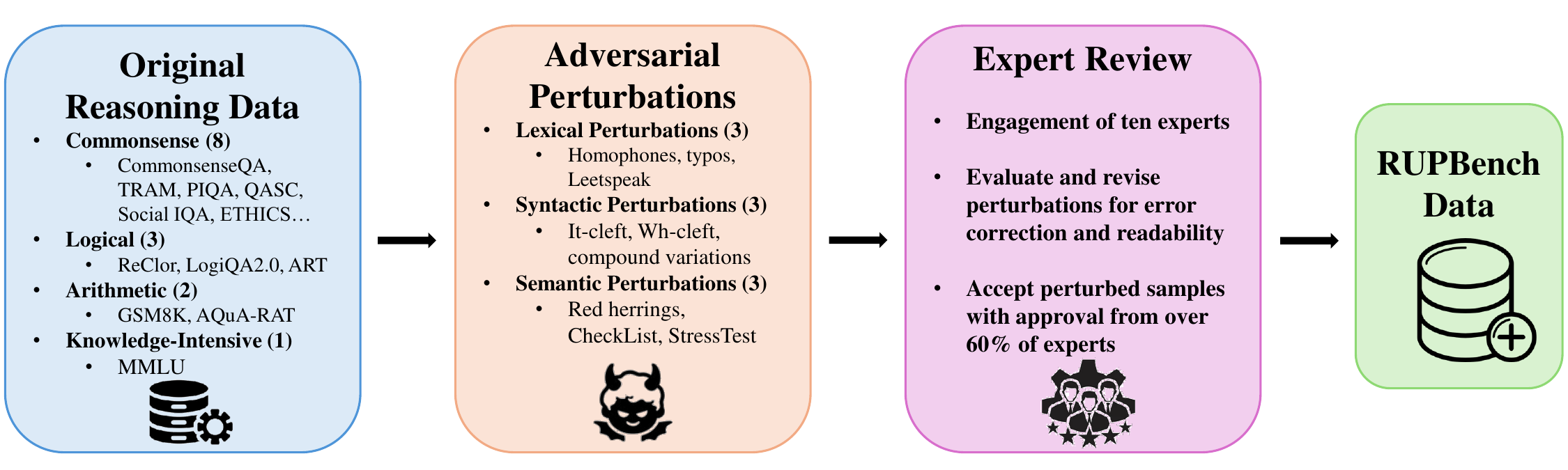}  
\caption{Overview of the data construction pipeline for RUPBench.}
\label{fig: data_pipeline}
\end{figure*}

\section{Related Work}
In this section, we provide an overview of LLM evaluation, with a focus on robustness. We also discuss the role of textual perturbations in assessing the robustness and safety of LLMs.
\subsection{LLM Evaluation}
Pretrained language models like BERT~\citep{kenton2019bert} and RoBERTa~\citep{liu2019roberta} have been the standard practice in many NLP tasks. However, the introduction of GPT-3~\citep{brown2020language} shifted the focus towards minimal fine-tuning approaches, such as zero-shot~\citep{kojima2022large} and few-shot learning. Recently, advanced LLMs like GPT-4o~\citep{gpt-4o}, Llama3~\citep{llama3modelcard}, and Gemini~\citep{team2023gemini} have demonstrated significant improvements across various domains, including complex reasoning~\citep{wang2023metacognitive, wang2023gemini, xia2024sportqa}, machine translation~\citep{ding2023towards}, and text classification~\citep{wang2022integrating, wang2023prominet}.

Given the remarkable performance of LLMs, their evaluation has garnered significant attention across areas like robustness~\citep{dong2023revisit}, hallucination~\citep{li2023halueval}, healthcare~\citep{wang2023large}, and ethics~\citep{wan2023kelly}. Benchmarks such as GLUE~\citep{wang2018glue} and SuperGLUE~\citep{wang2019superglue} have been foundational in advancing natural language understanding tasks. More recent benchmarks, including MMLU~\citep{hendrycks2020measuring}, BigBench~\citep{srivastava2023beyond}, and HellaSwag~\citep{zellers2019hellaswag}, assess capabilities in knowledge understanding and complex reasoning.

Robustness is particularly crucial for LLMs as it ensures reliable performance in diverse, real-world environments and the ability to handle noisy, incomplete, or adversarial inputs~\citep{wang2024empirical}. Existing benchmarks like AdvGLUE~\citep{wang2021adversarial} and AdvGLUE++~\citep{wang2024decodingtrust}, built on the foundation of GLUE, focus on evaluating robustness. However, these benchmarks do not sufficiently challenge the advanced capabilities of current LLMs, underscoring the need for more rigorous assessments.

Our benchmark, RUPBench, addresses this critical gap by incorporating diverse recent datasets that emphasize complex reasoning. This approach not only enhances performance differentiation but also pushes the boundaries of reasoning and knowledge in advanced LLMs, making it an essential tool for the next generation of LLM evaluation.

\subsection{Textual Perturbations and LLM Safety}
Textual perturbations involve creating variations in input text to evaluate the robustness and safety of LLMs. Unlike efforts aimed at generating potentially harmful outputs, such as SafetyPrompts~\citep{sun2023safety} or prompt injection attacks~\citep{esmradi2023comprehensive}, our perturbations mimic plausible user mistakes in data samples. Our goal is to ensure that LLMs can manage diverse, noisy, or slightly incorrect inputs without producing erroneous or harmful outputs, thereby enhancing their robustness and safety in real-world applications. Additionally, categorizing perturbations into lexical, syntactic, and semantic levels from a linguistic perspective covers a broad spectrum of text variations, enabling a nuanced understanding of how different perturbations affect LLM performance.

\section{Dataset Construction}
In this section, we introduce the 15 source reasoning datasets spanning commonsense, logic, arithmetic, and cross-domain areas. We describe the nine general text-based perturbations applied at lexical, syntactic, and semantic levels, resulting in a total of 365,580 perturbed samples. We also detail the involvement of human experts to ensure the quality and validity of the perturbations. The overall data construction pipeline is shown in Figure~\ref{fig: data_pipeline}.

\subsection{Tasks and datasets}

We consider 15 representative text-based reasoning datasets, which are categorized into four major reasoning groups: commonsense reasoning, arithmetic reasoning, logical reasoning, and knowledge-intensive reasoning. Table~\ref{tab: dataset_overview} provides an overview of the reasoning datasets and tasks.

\subsubsection{Commonsense Reasoning} This group encompasses nine datasets covering various dimensions of commonsense reasoning.
\begin{itemize}
\item \textbf{CommonsenseQA}~\citep{talmor2019commonsenseqa}: Focuses on general commonsense knowledge, requiring models to answer questions based on everyday scenarios.

\item \textbf{TRAM}~\citep{wang2023tram}: Assesses the model's ability to understand and reason about time-related information such as frequency, ordering, duration, and typical time.

\item \textbf{PIQA}~\citep{bisk2020piqa}: Targets physical interaction reasoning, challenging models with questions about everyday situations, favoring atypical solutions.

\item \textbf{QASC}~\citep{khot2020qasc}: Centers on scientific reasoning, requiring models to integrate and apply scientific knowledge to answer questions.

\item \textbf{Social IQA}~\citep{sap2019social}: Emphasizes social reasoning, evaluating the model's understanding of the social implications of everyday events and situations.

\item \textbf{Cosmos QA}~\citep{huang2019cosmos}: Focuses on contextual reasoning, requiring models to draw inferences from contextual information in narrative passages.

\item \textbf{NumerSense}~\citep{lin2020birds}: Tests numerical reasoning by requiring models to fill in missing numerical values (zero to ten) or  “no” in given sentences.

\item \textbf{RiddleSense}~\citep{lin2021riddlesense}: Challenges models to solve riddles that often require multiple pieces of commonsense knowledge and figurative language.

\item \textbf{ETHICS}~\citep{hendrycks2020aligning}: Focuses on moral reasoning, assessing the model's ability to make ethical judgments and understand moral principles.
\end{itemize}

\subsubsection{Arithmetic Reasoning} This group comprises two datasets focusing on math word problems.

\begin{itemize}
\item \textbf{GSM8K}~\citep{cobbe2021training}: Contains grade school math word problems requiring basic arithmetic and reasoning.

\item \textbf{AQuA-RAT}~\citep{ling2017program}: Comprises algebraic math word problems, requiring models to answer multiple-choice questions and generate rationales.
\end{itemize}

\subsubsection{Logical Reasoning} This group comprises three datasets focused on deductive reasoning (i.e., drawing conclusions based on premises) and abductive reasoning (i.e., forming hypotheses from incomplete information) tasks.

\begin{itemize}
\item \textbf{ReClor}~\citep{yu2019reclor}: Contains logical reasoning problems from standardized tests such as LSAT and GMAT, requiring models to perform deductive reasoning.

\item \textbf{LogiQA2.0}~\citep{liu2023logiqa}: Contains logical reasoning problems from the Chinese Civil Service Examination, including natural language inference (NLI) and machine reading comprehension (MRC) tasks.

\item \textbf{ART}~\citep{bhagavatula2019abductive}: Focuses on abductive reasoning, challenging models to select the most plausible explanation (hypothesis) given a pair of observations.
\end{itemize}

\subsubsection{Knowledge-Intensive Reasoning} We consider the MMLU~\citep{hendrycks2020measuring} benchmark as the standard for knowledge-intensive reasoning, encompassing a broad range of exam questions from 57 subjects across STEM, social sciences, humanities, and more.

\begin{table*}[ht]
\centering
\caption{Summary statistics of RUPBench. The benchmark is constructed using the validation or test sets from 15 source reasoning datasets, depending on availability and the presence of ground truth labels. ‘Pert.’ refers to perturbed, indicating the total number of samples after applying nine types of general perturbations to each original validation/test sample, with the original sample count shown in parentheses. For datasets like TRAM and ETHICS, which include multiple subtasks beyond commonsense reasoning, we extract the relevant samples for our analysis.}
\label{tab: dataset_overview}
\resizebox{\linewidth}{!}{%
\begin{tabular}{@{}lccccc@{}}
\toprule
\multirow{2}{*}{\textbf{Dataset}} & \multirow{2}{*}{\textbf{Domain}} & \multirow{2}{*}{\textbf{Answer Type}} & \textbf{\# Train Samples} & \textbf{\# Pert. Val/Test Samples} \\&  &  & \textit{(Source)}  & \textit{(RUPBench)}        \\ \midrule
\multicolumn{5}{c}{\textbf{Commonsense Reasoning}} \\ \midrule
CommonsenseQA  & General & 5-Way MC & 9,741 & 10,989 (1,221)\\ 
TRAM & Temporal  & 3-Way MC  & N/A & 29,610 (3,290) \\ 
PIQA & Physical & 2-Way MC  & 16,113 & 16,542 (1,838) \\ 
QASC & Science & 8-Way MC & 8,134 & 8,334 (926) \\ 
Social IQA & Social & 3-Way MC  & 33,410  & 17,586 (1,954) \\ 
Cosmos QA & Contextual & 4-Way MC & 25,262 & 26,865 (2,985) \\ 
NumerSense & Numerical & Number & 10,444 & 1,800 (200) \\ 
RiddleSense & Riddle & 5-Way MC & 3,510 & 9,189 (1,021) \\ 
ETHICS & Moral & 2-Way MC  & 13,910 & 35,676 (3,964)  \\ \midrule
\multicolumn{5}{c}{\textbf{Arithmetic Reasoning}} \\ \midrule
GSM8K & Grade School Math & Number & 7,473 & 11,871 (1,319) \\ 
AQuA-RAT & Algebra & 5-Way MC & 97,467 & 4,572 (508) \\ \midrule
\multicolumn{5}{c}{\textbf{Logical Reasoning}} \\ \midrule
ReClor & Deductive & 4-Way MC & 4,638 & 4,500 (500) \\ 
LogiQA2.0 & Deductive & 2/4-Way MC & 44,098 & 47,880 (5,320) \\ 
ART & Abductive & 2-Way MC & 169,654 & 13,788 (1,532 \\ \midrule
\multicolumn{5}{c}{\textbf{Knowledge-Intensive Reasoning}} \\ \midrule
MMLU & Multi-discipline & 4-Way MC & N/A & 126,378 (14,042) \\ 
\bottomrule
\end{tabular}}
\end{table*}

\subsection{Perturbation Categories}

We consider each reasoning dataset's validation or test sets as our source samples, upon which we perform various perturbations. Specifically, we categorize these perturbations into three major types: lexical, syntactic, and semantic. Our perturbations are designed to induce incorrect responses from the LLM while preserving the essence of the original content, ensuring that the ground truth answer remains unchanged despite the perturbations. Examples of RUPBench can be found in Appendix~\ref{data_examples}.

\subsubsection{Lexical Perturbation}
Lexical perturbations involve modifying individual words within the text to evaluate the model's robustness to variations. We consider three specific types of lexical perturbations: homophones, typos, and leetspeak, due to their ability to simulate common real-world challenges like phonetic confusion, typographical errors, and informal language.
\begin{itemize}
\item \textbf{Homophones}: This involves replacing words with their homophones, i.e., words that sound the same but have different meanings and spellings. For instance, “meet” might be replaced with “meat”.  Using the CMU Pronouncing Dictionary, we identify homophones for each word in a sentence and randomly select replacements.

\item \textbf{Typos}: This introduces random spelling errors into the text. Methods include swapping adjacent characters, inserting random characters, deleting characters, or replacing characters with random ones. For example, “example” might become “exmaple” or “ex@ample”.

\item \textbf{Leetspeak}~\citep{wei2024jailbroken}: This is a system of modified spellings used primarily on the Internet. This perturbation translates text into leetspeak by replacing letters with numbers or symbols that resemble them. For example, “write”  might become “WR1735”. Each character is mapped to a set of possible replacements, and one is randomly chosen.
\end{itemize}

\subsubsection{Syntactic Perturbation}
Syntactic perturbations involve modifying the structure of sentences to evaluate the model's understanding of grammar and sentence construction. We consider three specific types of syntactic perturbations: It-cleft, Wh-cleft, and compound variations. These perturbations are selected for their ability to challenge the model's syntactic parsing capabilities and emphasize different aspects of sentence structure and focus.

\begin{itemize}
\item \textbf{It-cleft}: This restructures sentences using the It-cleft construction, which highlights a specific part of the sentence by placing it after “It was”. For example, “The dog chased the cat” becomes “It was the dog that chased the cat”. This method involves using the spaCy library~\citep{honnibal2017spacy} to identify the subject, verb, and object in a sentence and rephrasing it to fit the It-cleft structure.

\item \textbf{Wh-cleft}: This restructures sentences using the Wh-cleft construction, which highlights a specific part of the sentence with Wh-words like “what”, “who”, “where”, etc. For example, “The dog chased the cat” becomes “What the dog chased was the cat”. Similar to the It-cleft method, we use the spaCy library to identify key elements and rephrase them to fit the Wh-cleft structure.

\item \textbf{Compound Variations}: This perturbation creates complex sentence structures by incorporating subordinating conjunctions, quantifiers, and modifying punctuation. For example, a simple sentence can be made more intricate with conjunctions like “although” and quantifiers like “several”. We use the NLTK library~\citep{bird2009natural} to tokenize sentences, identify parts of speech, and insert suitable conjunctions and quantifiers. Punctuation is then adjusted to form compound sentences.
\end{itemize}

\subsubsection{Semantic Perturbation}

Semantic perturbations modify the meaning or context of the text to evaluate the model's understanding of deeper linguistic aspects. We consider three specific types of semantic perturbations: Red herrings, CheckList~\citep{ribeiro2020beyond} items, and StressTest~\citep{naik2018stress} statements. These perturbations assess the model's ability to maintain logical consistency and focus on relevant information despite the presence of distracting, irrelevant, or misleading content.

\begin{itemize}
\item \textbf{Red Herrings (RHs)}: This introduces contextually plausible but irrelevant information designed to distract the model, aiming to challenge its focus on relevant parts of the text without altering the final answer. We use GPT-4o to generate these RHs, leveraging the efficiency and consistency of LLMs compared to human generation. We prompt GPT-4o with: “\textit{Given the statement: \{context\}, generate a single Red Herring either before, after, or within the original text to challenge the LLMs while keeping the original text and final answer intact}”. 

\item \textbf{CheckList}: This perturbation involves incorporating URLs, social media handles, or other irrelevant elements into the text. For example, embedding “@newswire” or “\url{http://dw.com}” within a sentence assesses the model's capability to manage such elements in context without being misled by their presence. We generate 100 random URLs and handles, with a subset selected to be inserted arbitrarily into various parts of each sample's context.

\item \textbf{StressTest}: This introduces logically redundant or repetitive phrases such as “and true is true”, “and false is not true”, or “if one is equal to one”. These phrases are inserted at random positions within the original text. The aim is to challenge models to maintain logical consistency and manage semantic redundancy.

\end{itemize}

\subsection{Expert Review}
After collecting the raw perturbed dataset, we conduct a human study involving ten human experts with at least an undergraduate degree to review the generated perturbations of each data sample, particularly the RHs, ensuring their quality and reliability. The experts evaluate whether the perturbations significantly alter the context or introduce errors that could mislead the models. If a perturbation is deemed unreadable, the experts rewrite it to align with the specific type. Their feedback is crucial for maintaining the original meaning of the text while effectively challenging the models. Any perturbations deemed implausible or overly disruptive are revised based on their insights. A perturbed data sample is considered acceptable without further changes if it receives approval from at least 60\% of the experts (i.e., six out of ten).

\begin{table*}[ht]
\centering
\caption{Model performance on RUPBench, including raw and perturbed datasets. The results are averaged over three runs. The numbers outside parentheses represent the accuracy (\%) on the original data, while the numbers within parentheses indicate the average PDR (\%) across nine perturbations.}
\label{tab:combined}
\resizebox{\textwidth}{!}{
\begin{tabular}{@{}lccccccc@{}}
\toprule
 \multirow{2}{*}{\textbf{Dataset}} & \textbf{Gemma} & \textbf{Phi-3-mini} & \textbf{Gemma} & \textbf{Llama3 } & \textbf{Phi-3-medium} & \textbf{Llama3} & \textbf{GPT-4o} \\ & \textit{2B} & \textit{3.8B} & \textit{7B} & \textit{8B} & \textit{14B} & \textit{70B} & \textit{>175B} \\
\midrule
CommonsenseQA & 45.6 (28.5) & 75.8 (24.7) & 66.0 (24.1) & 73.5 (11.3) & 80.3 (18.4) & 80.7 (12.4) & 83.9 (5.5) \\
TRAM & 53.6 (20.2) & 79.4 (9.5) & 67.3 (21.1) & 78.8 (6.1) & 81.3 (10.6) & 82.8 (8.5) & 87.8 (7.8)\\
PIQA & 50.1 (1.1) & 79.5 (0.6) & 73.3 (0.3) & 81.3 (1.2) & 83.7 (0.9) & 82.1 (0.7) & 91.2 (0.5) \\
QASC & 61.4 (39.0) & 77.3 (18.4) & 67.1 (35.4) & 75.9 (17.3) & 75.3 (20.7) & 79.6 (16.9) & 92.6 (14.5) \\
Social IQA & 53.1 (8.7) & 70.3 (3.5) & 62.1 (5.3) & 70.4 (5.5) & 73.8 (6.2) & 74.1 (8.3) & 80.7 (8.8) \\
Cosmos QA & 52.4 (2.2) & 72.7 (5.6) & 64.0 (0.9) & 81.2 (3.6) & 82.9 (4.2) & 86.1 (6.5) & 88.6 (3.6)\\
NumerSense & 37.8 (86.3) & 66.4 (93.9) & 62.5 (53.3) & 64.8 (15.8) & 68.2 (84.3) & 69.5 (18.9) & 83.2 (20.8)\\
RiddleSense & 37.1 (24.9) & 58.5 (22.2) & 50.8 (20.9) &  64.1 (17.3) & 63.3 (20.3) & 70.7 (18.4) & 89.3 (16.7)\\ 
ETHICS & 40.8 (13.3) & 56.0 (7.7) &  61.7 (10.3) & 78.1 (12.3) & 69.2 (6.8) & 82.3 (11.8) & 94.7 (7.8)\\
\midrule
GSM8K & 16.4 (49.8) & 70.3 (22.2) & 45.6 (40.5) & 76.7 (18.2) & 81.2 (26.7) & 85.9 (20.3) & 94.1 (22.5) \\
AQuA-RAT & 19.6 (-0.3) & 26.1 (6.2) & 30.1 (-2.0) & 38.7 (17.6) & 32.8 (9.8) & 41.5 (19.2) & 48.2 (12.3) \\
\midrule
ReClor & 32.1 (10.4) & 62.0 (8.4) & 41.9 (9.3) & 63.1 (9.0) & 67.9 (13.2) & 69.5 (12.5) & 77.2 (8.9)\\
LogiQA2.0 & 42.8 (6.3) & 55.9 (5.9) & 51.4 (3.7) & 55.7 (5.5) & 58.3 (5.7) & 60.4 (7.0) & 72.8 (6.6) \\
ART & 57.3 (9.4) & 78.3 (8.8) & 68.8 (2.2) & 73.6 (1.1) & 79.8 (10.3) & 80.2 (1.8) & 87.1 (3.7) \\
\midrule
MMLU & 40.5 (18.9) & 63.8 (6.3) & 62.5 (15.2) & 67.3 (7.7) & 76.8 (7.2) & 80.2 (9.3) & 87.6 (9.7)\\
\midrule
\textbf{Average} & 42.7 (21.2) & 66.1 (16.3) & 58.3 (16.0) & 69.5 (10.0) & 71.6 (16.3) & 75.0 (11.5) & 83.9 (10.0) \\
\bottomrule
\end{tabular}}
\end{table*}

\section{Experiments}
In this section, we describe the experimental setup, report overall performance, analyze robustness from different perspectives, and perform error analysis to identify common errors in LLMs under original and perturbed texts.
\subsection{Experimental Setup}
We evaluate several leading LLMs for RUPBench on original and perturbed samples, including GPT-4o~\citep{gpt-4o}, Llama3-8B-Instruct, Llama3-70B-Instruct~\citep{llama3modelcard}, Phi-3-mini-128k-Instruct, Phi-3-medium-128k-Instruct~\citep{abdin2024phi}, Gemma-2B-Instruct, and Gemma-7B-Instruct~\citep{team2024gemma}. GPT-4o is accessed through the OpenAI API, while the other models are loaded from Hugging Face. For generating model responses, we use greedy decoding (temperature = 0). Due to API cost constraints, we randomly sample 300 instances per dataset (except NumerSense), each with 10 variations (one raw and nine perturbed). For MMLU, we sample 50 instances per subject. We utilize 5-shot Chain-of-Thought prompting~\citep{kojima2022large} for arithmetic reasoning datasets, while applying 5-shot standard prompting for the other datasets.

For evaluation metrics, we report the original performance using accuracy, suitable for the multiple-choice nature of most tasks. Additionally, following~\citep{zhu2023promptbench}, we report the Performance Drop Rate (PDR) to measure the relative performance decline after adversarial perturbations. A negative PDR indicates instances where perturbations can unexpectedly improve performance.

\subsection{Results and Analysis}
We compare the performance of multiple LLMs across all datasets, followed by a robustness analysis considering perturbation types, task types, and models. Finally, we conduct an error analysis to identify LLM weaknesses under perturbations.

\subsubsection{Main Results}
We present the overall performance of various models on RUPBench reasoning datasets, comparing original and perturbed samples. GPT-4o demonstrates the highest accuracy with an average of 83.9\% and the lowest average PDR of 10.0\%, indicating its strong robustness to adversarial perturbations. Among the open-source LLMs, Llama3-70B performs exceptionally well with a relatively low PDR of 11.5\%. In contrast, the smallest model, Gemma-2B, shows the lowest average accuracy of 42.7\% and the highest PDR of 21.2\%, highlighting its susceptibility to perturbations.

In terms of datasets, CommonsenseQA presents notable variability. Gemma-2B achieves only 45.6\% accuracy with a substantial PDR of 28.5\%, whereas GPT-4o reaches 83.9\% accuracy with a significantly lower PDR of 5.5\%. This trend is consistent across most datasets, where larger models generally perform better and exhibit greater robustness. For instance, in the GSM8K dataset, GPT-4o achieves 94.1\% accuracy with a PDR of 22.5\%, compared to Gemma-2B’s 16.4\% accuracy and 49.8\% PDR.

Interestingly, models demonstrate varied responses to specific perturbations. The arithmetic reasoning datasets GSM8K and AQuA-RAT show mixed results, with AQuA-RAT experiencing negative PDRs for some models, such as -0.3\% for Gemma-2B and -2.0\% for Gemma-7B, suggesting that certain perturbations might inadvertently aid performance in these tasks.

Overall, while the largest models like GPT-4o exhibit robust performance with minimal PDRs, smaller models like Gemma-2B and Phi-3-mini-3.8B struggle significantly more in challenging datasets like NumerSense and GSM8K. This underscores the necessity for further advancements in model robustness and the importance of evaluating models on diverse and complex reasoning tasks.

\subsubsection{Robustness Analysis}
We investigate robustness across nine perturbation types within three major categories (lexical, syntactic, and semantic) and the relationship between the robustness of reasoning data types and models.

\begin{figure}[htbp]
\centering
\includegraphics[width=.5\textwidth]{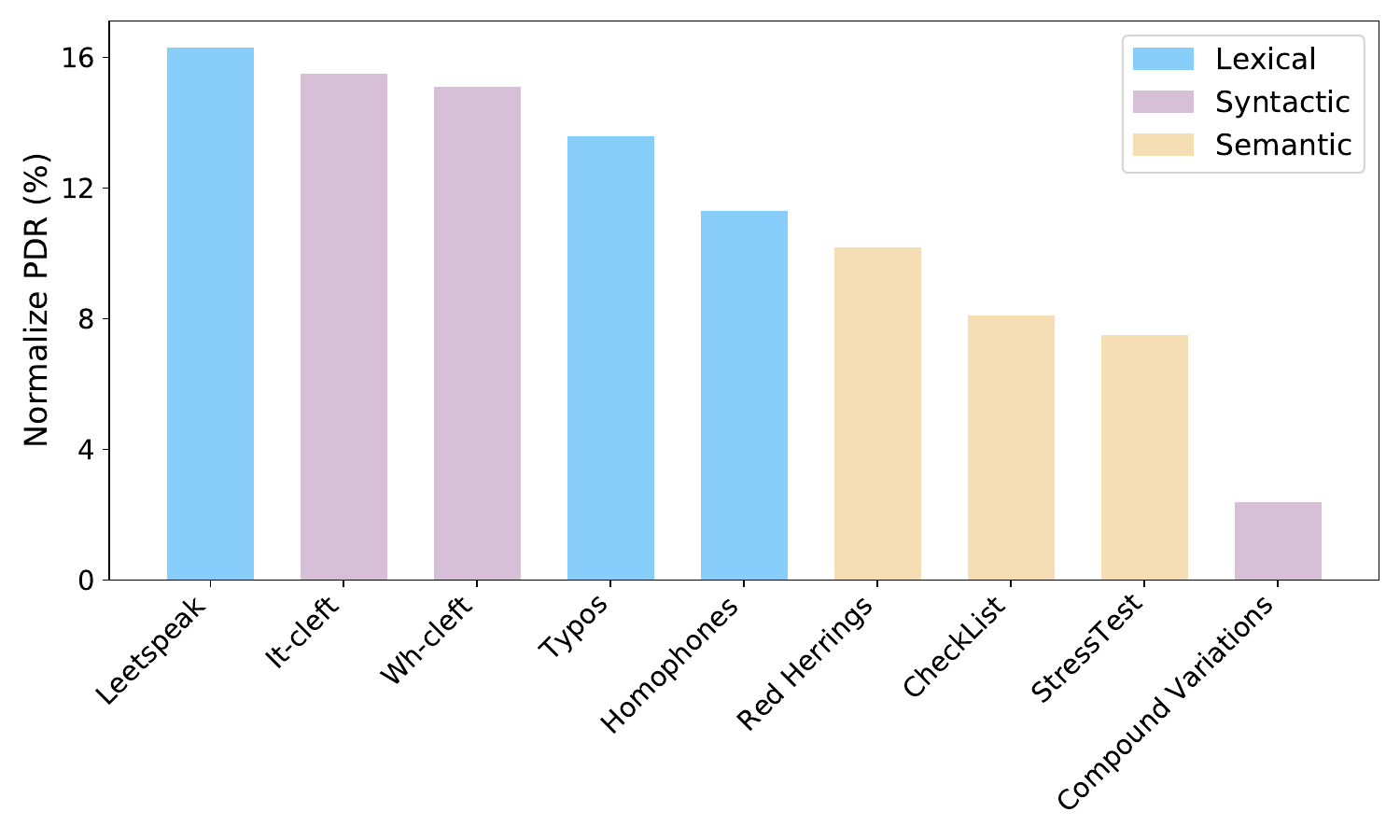}  
\caption{Normalized PDR (\%) of nine perturbation types, averaged across datasets and models. Normalization scales each perturbation's impact.}
\label{fig: robustness_perturbation}
\end{figure}

\noindent \textbf{Perturbation Categories}
Figure~\ref{fig: robustness_perturbation} displays the normalized PDR (measure for robustness) for nine perturbation types, averaged across datasets and models. Lexical perturbations, particularly Leetspeak (16.3\%) and typos (13.6\%) result in high PDRs, likely due to the models' reliance on precise word forms and spelling to understand context and meaning, making them highly sensitive to such variations. Syntactic perturbations, especially It-cleft (15.5\%) and Wh-cleft (15.1\%) constructions, also cause significant performance drops. Models may struggle with non-standard sentence structures that deviate from the syntactic patterns they are trained on, potentially confusing their parsing mechanisms and affecting comprehension. Finally, semantic perturbations like Red Herrings (10.2\%) exhibit notable PDRs, indicating that introducing irrelevant information can distract and mislead the models.

\begin{figure*}[htbp]
\centering
\includegraphics[width=\textwidth]{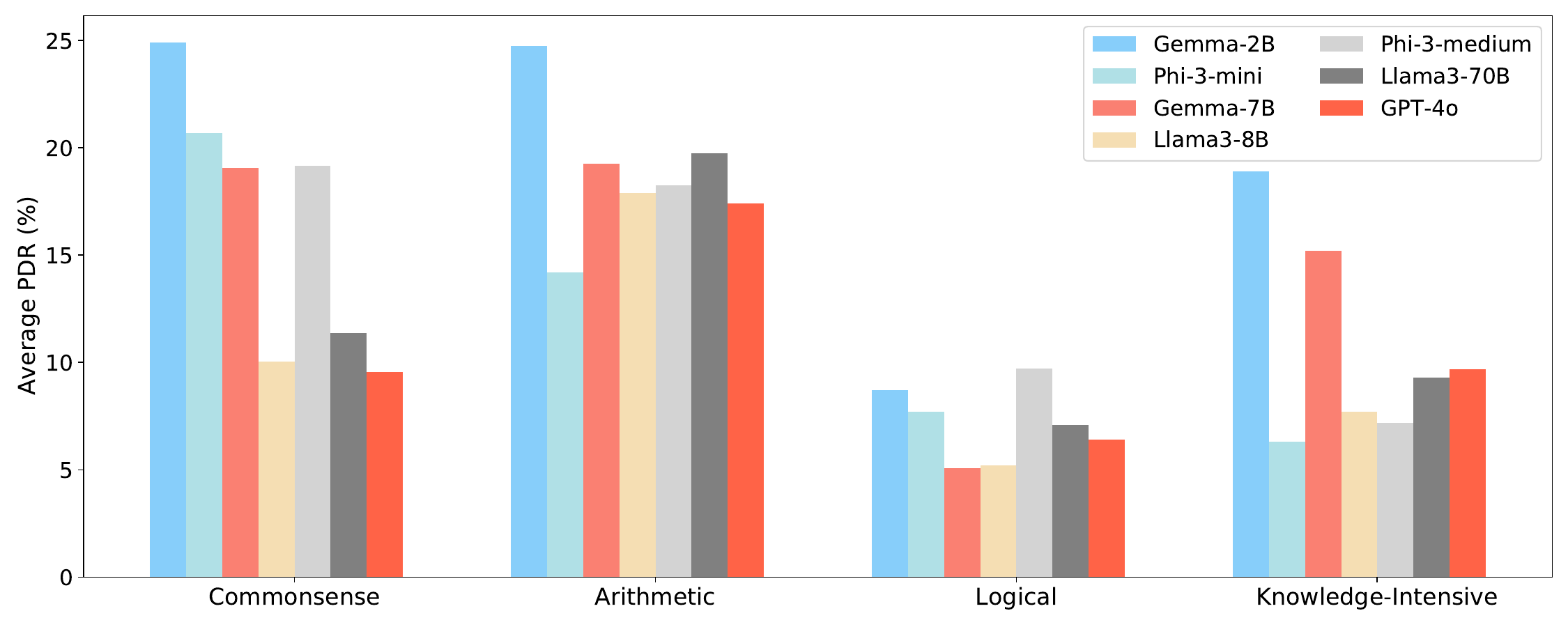}  
\caption{Average PDR (\%) by dataset categories and models. Each bar represents the average PDR for a specific model across different dataset categories. Commonsense reasoning and arithmetic reasoning are generally more susceptible to perturbations. Additionally, larger models tend to be more robust to perturbations.}
\label{fig: robustness_data_model}
\end{figure*}

\noindent \textbf{Data Categories and Models} 
We further examine the impact of data categories and models on robustness through average PDR, as shown in Figure~\ref{fig: robustness_data_model}. The results demonstrate that the small-size LLM Gemma-2B is more susceptible to perturbations compared to the other LLMs. As model size increases, there is a general trend towards improved robustness, indicated by a decrease in PDR. Commonsense and arithmetic reasoning tasks are more affected by perturbations, as evidenced by their higher PDRs. This can be attributed to these tasks' reliance on specific contextual knowledge and precise calculations, which are more easily disrupted. Conversely, logical and knowledge-intensive reasoning tasks exhibit lower PDRs, likely due to their structured nature and extensive training data, making them more resilient to perturbations.

\subsubsection{Error Analysis}
We provide a detailed examination of the errors encountered by LLMs. Through manual inspection of incorrect predictions under perturbations, we find that in commonsense reasoning, errors often involve context misinterpretation (32.7\%) and literal interpretation (28.2\%), exacerbated by perturbations that introduce ambiguities or misleading details. In arithmetic reasoning, the most common mistakes are calculation errors (35.9\%) and misunderstandings of word problems (28.4\%), amplified by perturbations that alter problem wording. Logical reasoning errors typically include faulty deductions (30.7\%) and inconsistent reasoning (27.0\%), often due to syntactic perturbations that disrupt the logical flow. In knowledge-intensive reasoning, the primary issues are knowledge gaps (40.3\%) and concept confusion (26.9\%), with semantic perturbations introducing irrelevant or contradictory information that challenges the model's knowledge base. This analysis highlights specific challenges posed by different perturbation types, emphasizing the need for targeted strategies to enhance LLM robustness. More details on each error type and their proportions under different reasoning tasks can be found in Appendix~\ref{error_types}.

\section{Discussion}
Investigating robustness is essential for ensuring the reliable use of LLMs. In this work, we introduce RUPBench, a comprehensive benchmark that incorporates 15 reasoning datasets with nine general perturbations, covering lexical, syntactic, and semantic challenges for evaluating LLM robustness. Our study reveals significant variability in the robustness of different LLMs across various reasoning tasks. Generally, larger models tend to be less susceptible to perturbations. Additionally, LLMs are more vulnerable to lexical and syntactic perturbations. They exhibit varying levels of resilience across different types of reasoning tasks, highlighting the influence of data nature on model robustness. Finally, we identify error patterns that help understand the inherent weaknesses in LLMs and provide direction for targeted improvements.

For future work, we will incorporate more challenging and diverse perturbation types to simulate real-world adversarial inputs. Additionally, integrating domain-specific datasets and perturbations can provide deeper insights into model performance in specialized fields such as healthcare, legal, and finance. Finally, we will continuously update RUPBench with emerging datasets and perturbations to ensure rigorous LLM robustness evaluation for the community.

\section{Limitations}
We acknowledge several limitations in our study. First, our evaluation is performed on a subset of data samples, which may not fully capture the comprehensive robustness of LLMs. Second, although our benchmark includes diverse datasets, perturbations, and models, it is impractical to encompass all possible LLMs, datasets, and adversarial perturbations due to computational constraints. Third, we do not explore sufficient prompting methods, which can be crucial for assessing LLMs' the general and robustness performance. Lastly, our use of textual questions may not entirely reflect the robustness capabilities of LLMs, as real-world scenarios often involve multimodal cues such as images and videos. Future research could extend similar evaluation pipelines to multimodal LLMs to provide a more comprehensive assessment.

\bibliography{custom}

\appendix

\newpage
\section{RUPBench Examples}
\label{data_examples}
We present RUPBench examples with nine perturbation types, covering lexical, syntactic, and semantic-level changes, in Table~\ref{tab:data_sample}.

\begin{table*}[ht]
\centering
\caption{Examples of RUPBench for each perturbation type. OS (Original Sentence) and PS (Perturbed Sentence) are presented, with major changes highlighted in red and blue.}
\label{tab:data_sample}
\resizebox{\textwidth}{!}{
\begin{tabular}{ccl}
\hline
\textbf{Data} & \textbf{Perturbation} & \textbf{Sample} \\
\hline
CommonsenseQA & Homophone & \begin{tabular}[c]{@{}l@{}}OS: Where \textcolor{blue}{do apples} form on an \textcolor{blue}{apple} tree? \\ PS:  Where \textcolor{red}{deux apple's} form on an \textcolor{red}{appel} tree?\end{tabular} \\ \hline
PIQA & Typo & \begin{tabular}[c]{@{}l@{}}OS: How \textcolor{blue}{to finish wood} table after pictures have \textcolor{blue}{been glued} on.\\ PS: How \textcolor{red}{tV funish womod} table after pictures have \textcolor{red}{beedn gOlued} on.\end{tabular} \\ \hline
Social IQA & Leetspeak & \begin{tabular}[c]{@{}l@{}}
OS: Robin had been away for two weeks \textcolor{blue}{on} his honeymoon. \\
Cameron \textcolor{blue}{picked} him up \textcolor{blue}{on} his return home. \\
PS: Robin had been away for two weeks \textcolor{red}{0/\/} his honeymoon. \\
Cameron \textcolor{red}{|D!<|<â‚¬|)} him up \textcolor{red}{()/\/} his return home.\end{tabular} \\ \hline
TRAM & It-cleft & \begin{tabular}[c]{@{}l@{}}
OS: \textcolor{blue}{Several tenants blame other neighbors as perpetrators} \\ of the rift, however. How long has there been a rift between neighbors? \\
PS: \textcolor{red}{It was several tenants who blame other neighbors as perpetrators} \\ of the rift, however.
How long has there been a rift between neighbors? \\ \end{tabular} \\ \hline
ART & Wh-cleft & \begin{tabular}[c]{@{}l@{}}OS: \textcolor{blue}{Anna was} making a world atlas. \textcolor{blue}{Then she colored} in her atlas.\\ PS: \textcolor{red}{What Anna was doing was} making a world atlas. \\ \textcolor{red}{What she did next was color} in her atlas. \end{tabular} \\ \hline
RiddleSense & \makecell{Compound \\ Variation} & \begin{tabular}[c]{@{}l@{}}OS: What is always slow to come\textcolor{blue}{,} but never actually happens\textcolor{blue}{?} \\ PS:  \textcolor{red}{If} What is always slow to come \textcolor{red}{, , }but never actually happens \textcolor{red}{ ?}\end{tabular} \\ 
\hline
GSM8K & Red Herrings & \begin{tabular}[c]{@{}l@{}}
OS: James delivers 600 newspapers in a day. He delivers 198 newspapers \\
to District A, some to District B and 209 newspapers to District C. \\
How many newspapers does he deliver to District B? \\
PS: James, \textcolor{red}{who wakes up at 5 am every morning}, delivers 600 newspapers in a day. \\
He delivers 198 newspapers to District A, some to District B, \\ and 209 newspapers to District C. \textcolor{red}{On Sundays, he also delivers a special magazine} \\ \textcolor{red}{to each house.} How many newspapers does he deliver to District B? \end{tabular} \\ \hline
NumerSense & CheckList & \begin{tabular}[c]{@{}l@{}}OS: boeing and lockheed are <mask> aeronautics companies. \\ PS: \textcolor{red}{\$https://github.com\$ \$http://huffpost.com\$} boeing \\ \textcolor{red}{\$https://medium.com/\@writer\$ \$http://huffpost.com\$} and \\ \textcolor{red}{\$\@tech\_updates\$} lockheed are <mask> aeronautics companies.\end{tabular} \\ \hline
QASC & StressTest & \begin{tabular}[c]{@{}l@{}}
OS: Breaking complex chemicals into simple ones in humans occur in what location? \\
PS: Breaking complex chemicals into simple ones in humans occur in what location? \\ \textcolor{red}{and false is not true and fire is hot and the sky is blue} \\ \textcolor{red}{if gravity pulls objects down if one is equal to one.} \end{tabular} \\ \hline
\end{tabular}}
\end{table*}

\section{Error Types}
\label{error_types}
Table~\ref{tab:common_errors} illustrates the major error types in LLMs for different reasoning tasks under perturbations.

\begin{table}[ht]
\centering
\caption{Distribution of major error types in LLMs by reasoning tasks under perturbations. Con. Misinter. refers to context misinterpretation, and Misunder. refers to misunderstanding.}
\label{tab:common_errors}
\resizebox{.5\textwidth}{!}{
\begin{tabular}{llc}
\hline
\textbf{Task} & \textbf{Error Types} & \textbf{Proportion (\%)} \\
\hline
\multirow{4}{*}{Commonsense}
& Con. Misinter. & 32.7 \\
& Literalism & 28.2 \\
& Surface Patterns & 23.8 \\
& Ignored Details & 15.3 \\
\hline
\multirow{4}{*}{Arithmetic}
& Calculation Mistakes & 35.9 \\
& Word Misunder. & 28.4 \\
& Logical Steps & 25.8 \\
& Unit Errors & 9.9\\
\hline
\multirow{4}{*}{Logical}
& Faulty Deduction & 30.7 \\
& Inconsistency & 27.0 \\
& Wrong Assumptions & 23.9 \\
& Connective Misuse & 18.4 \\
\hline
\multirow{4}{*}{\makecell{Knowledge-\\Intensive}}
& Knowledge Gaps & 40.3 \\
& Concept Confusion & 26.9 \\
& Fact Errors & 21.3 \\
& Data Misuse & 11.5 \\
\hline
\end{tabular}}
\end{table}

For commonsense reasoning tasks, errors often include context misinterpretation (32.7\%), where the model fails to grasp the overall context, leading to incorrect conclusions. For example, given the statement “John went to the bank to deposit his paycheck”, the model might incorrectly assume “bank” refers to the side of a river rather than a financial institution. Literalism (28.2\%) is another common error, where the model interprets idiomatic or figurative language literally, resulting in incorrect responses. An example is misinterpreting “kick the bucket” as physically kicking a bucket instead of understanding it as an idiom for dying. Additionally, reliance on surface patterns (23.8\%) occurs when the model focuses on superficial text features rather than underlying meanings, such as recognizing “dog" and “bark" but failing to understand that “bark” refers to the sound made by a dog. Ignored details (15.3\%) represent instances where the model overlooks crucial information, significantly impacting the answer. For instance, it might miss the importance of “only” in “She only eats vegetables” leading to incorrect dietary assumptions.

In arithmetic reasoning, calculation mistakes (35.9\%) are the most frequent errors, where the model makes errors in mathematical computations, such as adding 5 + 7 and incorrectly arriving at 11. Word misunderstandings (28.4\%) occur when the model misinterprets the problem's wording, leading to incorrect problem-solving steps. For example, it might misinterpret “double” in “double the number” as simply repeating the number rather than multiplying it by two. Errors in logical steps (25.8\%) involve incorrect or missing steps in the solution process, such as skipping a step in a multi-step algebra problem. Unit errors (9.9\%) arise when the model confuses or mishandles units of measurement, such as mixing up centimeters and inches, affecting the accuracy of the solution.

For logical reasoning tasks, faulty deduction (30.7\%) is a common error, where the model draws incorrect conclusions from the given premises due to flawed reasoning. For instance, given “All humans are mortal. Socrates is a human”, the model might incorrectly conclude that “Socrates is not mortal”. Inconsistency (27.0\%) occurs when the model's reasoning is not logically coherent, such as providing contradictory answers to similar questions. Wrong assumptions (23.9\%) involve the model making incorrect initial assumptions that lead to errors in the logical process, like assuming all birds can fly when solving a problem about penguins. Connective misuse (18.4\%) refers to incorrect use of logical connectors, such as misinterpreting “if” and “only if”, which disrupts the logical flow of the argument.

In knowledge-intensive reasoning, the primary issue is knowledge gaps (40.3\%), where the model lacks the necessary background information to answer correctly, indicating limitations in the model's training data. For instance, it might not know that “Einstein developed the theory of relativity". Concept confusion (26.9\%) involves the model mixing up related but distinct concepts, leading to incorrect answers, such as confusing “mitosis” and “meiosis” in a biology question. Fact errors (21.3\%) occur when the model recalls or generates incorrect factual information, like stating that “Albert Einstein won the Nobel Prize in Chemistry for his discovery of the photoelectric effect”. Data misuse (11.5\%) happens when the model incorrectly applies relevant data, leading to erroneous conclusions, such as using outdated statistics to answer a current events question, highlighting challenges in the model's data integration capabilities.

\section{Datasheet}
We provide the datasheet for RUPBench following~\citep{gebru2021datasheets}.

\begin{center}
\fbox{\textbf{OVERVIEW}}
\end{center}

\noindent \textbf{Motivation and Intended Uses.}

\noindent {\color{violet}{1. What are the intended purposes for this benchmark?}}

\noindent The intended purposes of RUPBench are to systematically evaluate the robustness of large language models (LLMs) across a diverse set of reasoning tasks and to provide insights into their performance under various types of textual perturbations. By offering a comprehensive benchmark, RUPBench aims to help researchers and practitioners identify and address specific weaknesses in LLMs, thereby enhancing their reliability and effectiveness in real-world applications.

\noindent {\color{violet}{2. Was it designed to address a specific task or fill a particular gap in research or application?}}

\noindent Yes, RUPBench was specifically designed to fill a gap in the evaluation of LLMs' robustness. While existing benchmarks often focus on restricted tasks or types of perturbations, RUPBench provides a more holistic framework that encompasses a wide range of reasoning tasks (commonsense, arithmetic, logical, and knowledge-intensive) and three major categories of textual perturbations (lexical, syntactic, and semantic). This allows for a more nuanced understanding of how LLMs perform under various adversarial conditions, addressing the need for a rigorous and multifaceted robustness evaluation.

\vspace{5mm}

\noindent \textbf{Limitations and Inappropriate Uses.}

\noindent {\color{violet}{3. Are there any specific tasks or applications for which this benchmark should not be used?}}

\noindent RUPBench is specifically designed to evaluate the robustness of LLMs in reasoning tasks under various textual perturbations. It is not suitable for tasks such as natural language generation, summarization, or translation. Additionally, it is not designed for evaluating LLMs in highly specialized or domain-specific applications outside the scope of the included datasets, such as biomedical text analysis or highly technical legal document processing, unless those fields are represented in the included datasets and perturbations. The benchmark is also not intended for use in evaluating non-textual data or multimodal tasks that combine text with other data types, such as images or audio.

\begin{center}
\fbox{\textbf{DETAILS}}
\end{center}

\noindent \textbf{Composition.}

\noindent {\color{violet}{4. What do the instances that comprise the benchmark represent?}}

\noindent The instances in RUPBench represent various reasoning tasks, specifically designed to test the robustness of LLMs. Each instance includes a reasoning question or problem from one of the four major categories: commonsense (CommonsenseQA, TRAM, PIQA, QASC, Social IQA, Cosmos QA, NumerSense, RiddleSense, ETHICS), arithmetic (GSM8K, AQuA-RAT), logical (ReClor, LogiQA2.0, ART), and knowledge-intensive (MMLU) reasoning. These instances are further subjected to nine types of textual perturbations, covering lexical (homophones, typos, Leetspeak), syntactic (It-cleft, Wh-cleft, compound variation), and semantic levels (red herrings, CheckList, StressTest), to simulate real-world input variations and assess how well LLMs handle such adversarial conditions.

\noindent {\color{violet}{5. How many instances are there in total (of each type, if appropriate)?}}

\noindent RUPBench consists of a total of 365,580 instances (excluding the original instances). This includes 15 original datasets, each subjected to nine different types of perturbations. Specifically, the number of perturbed samples for each dataset is as follows: CommonsenseQA (10,989), TRAM (29,610), PIQA (16,542), QASC (8,334), Social IQA (17,586), Cosmos QA (26,865), NumerSense (1,800), RiddleSense (9,189), ETHICS (36,676), GSM8K (11,871), AQuA-RAT (4,572), ReClor (4,500), LogiQA2.0 (47,800), ART (13,788), and MMLU (126,378).

\noindent {\color{violet}{6. Does the benchmark contain all possible instances or is it a sample (not necessarily random) of instances from a larger set?}}

\noindent The benchmark contains a curated selection of instances from the available reasoning datasets, specifically from the validation or test sets.

\noindent {\color{violet}{7. Is there a label or target associated with each instance?}}

\noindent Yes, each instance in the benchmark has an associated label or target. These labels represent the correct answers or expected outputs for the reasoning tasks, which are used to evaluate the performance and robustness of the LLMs.

\noindent {\color{violet}{8. Is the benchmark self-contained, or does it link to or otherwise rely on external resources (e.g., websites, tweets, other datasets)?}}

\noindent RUPBench is built upon existing datasets but is self-contained. It includes perturbed versions of instances from various established reasoning datasets. While the original datasets are sourced from external resources, RUPBench itself provides all necessary data for robustness evaluation without requiring access to the external sources. Users do not need to access the original datasets separately, as all relevant instances and their perturbations are included within RUPBench.

\noindent {\color{violet}{9. Does the benchmark contain data that might be considered sensitive in any way?}}

\noindent The benchmark does not contain any sensitive data.

\vspace{5mm}
\noindent \textbf{Data Quality.}

\noindent {\color{violet}{10. Is there any missing information in the benchmark?}}

\noindent Everything is included. No data is missing.

\noindent {\color{violet}{11. What errors, sources of noise, or redundancies are important for benchmark users to be aware of?}}

\noindent Benchmark users should be aware of potential sources of noise and errors, such as inconsistencies in how perturbations are applied to different instances, which may affect model performance evaluation. Some perturbations may introduce subtle ambiguities or irrelevant information that could disproportionately impact certain types of reasoning tasks, leading to variability in results. Additionally, redundancies might arise if multiple perturbations affect the same aspect of an instance, potentially skewing the analysis. It's also important to consider that manual inspection and correction of perturbations, while thorough, may still leave room for subjective interpretations, which could introduce a level of bias into the benchmark.

\noindent {\color{violet}{12. How was the data validated/verified?}}

\noindent The data in RUPBench was validated and verified through a multi-step process. First, each source dataset underwent a thorough review through sampling instances to ensure quality. Perturbations were then generated and applied to these instances following standardized procedures to maintain consistency across the benchmark.

To ensure the quality and reliability of the perturbed data, a human study was conducted involving ten experts with at least an undergraduate degree. These experts reviewed the generated perturbations to verify that they maintained human readability while introducing the intended adversarial variations. If a perturbation was deemed unreadable or significantly altered the context, the experts would rewrite it to align with the specific perturbation type.

Finally, any identified errors or inconsistencies were corrected based on expert feedback, and a consensus approach was used to ensure that at least 60\% of experts approved each perturbed instance.

\vspace{5mm}
\noindent \textbf{Pre-Processing, Cleaning, and Labeling.}

\noindent {\color{violet}{13. What pre-processing, cleaning, and/or labeling was done on this benchmark?}}

\noindent Original datasets underwent human reviews for quality checks. Nine types of textual perturbations were systematically applied to each dataset, covering lexical, syntactic, and semantic levels. These perturbations were designed to simulate real-world input variations and test the robustness of the models. In particular, for the arithmetic reasoning datasets GSM8K and AQuA-RAT, no numerical alterations were made to keep the final answers unchanged. Finally, the perturbed samples were reviewed by a panel of ten experts to ensure the perturbations maintained readability and did not introduce significant context alterations. Experts corrected any perturbations that were unreadable or inappropriate.

\noindent {\color{violet}{14. Provide a link to the code used to pre-process/clean/label the data, if available.}}

\noindent The code for data pre-processing is available on the official GitHub page.

\noindent {\color{violet}{15. If there are any recommended data splits (e.g., training, validation, testing), please explain.}}

\noindent RUPBench is designed primarily for the evaluation of LLM robustness and does not include predefined splits for training, validation, or testing. Instead, it provides a comprehensive set of perturbed instances intended for testing the performance of already trained models. Users are encouraged to use the entire dataset for evaluation purposes. If specific splits are required for custom analyses or experiments, users can create their own training, validation, and testing splits as appropriate for their specific needs. Alternatively, users can use the training set of the source dataset, if available, and validate the test samples in RUPBench.

\vspace{2mm}
\fbox{
  \begin{minipage}{.43\textwidth}
    \centering
    \textbf{ADDITIONAL DETAILS ON} \\
    \textbf{DISTRIBUTION AND MAINTENANCE}
  \end{minipage}
}

\vspace{2mm}
\noindent \textbf{Distribution.}

\noindent {\color{violet}{16. Will the benchmark be distributed to third parties outside of the entity (e.g., company, institution, organization) on behalf of which the dataset was created?}}

\noindent Yes, the benchmark will be publicly available on the Internet.

\noindent {\color{violet}{17. How will the benchmark be distributed (e.g., tarball on website, API, GitHub)?}}

\noindent The benchmark is distributed via the official GitHub page.

\noindent {\color{violet}{18. When will the benchmark be distributed?}}

\noindent The benchmark will be released in June 2024.

\vspace{5mm}
\noindent \textbf{Maintenance.}

\noindent {\color{violet}{19. Who will be supporting/hosting/maintaining the benchmark?}}

\noindent The first author of the RUPBench paper will support and maintain the benchmark. 

\noindent {\color{violet}{20. Will the benchmark be updated (e.g., to correct labeling errors, add new instances, delete instances)?}}

\noindent Updates to test sets and error corrections will be shared on the official GitHub page.

\noindent {\color{violet}{21. Will older versions of the benchmark continue to be supported/hosted/maintained?}}

\noindent Given any updates to the benchmark, older versions will be retained for consistency.

\noindent {\color{violet}{22. If others want to extend/augment/build on/contribute to the benchmark, is there a mechanism for them to do so?}}

\noindent Anyone interested in incorporating fixes or extensions should reach out to the original authors of RUPBench.

\end{document}